\documentclass[11pt,twoside]{article}

\usepackage{asp2014}

\pdfoutput=1

\aspSuppressVolSlug
\resetcounters

\begin{document}

\title{Galaxy Image Simulation Using Progressive GANs}

\author{Mohamad~Dia,$^1$ Elodie ~Savary,$^2$ Martin~Melchior,$^1$ and Frederic~Courbin$^2$}
\affil{$^1$Institute for Data Science, University of Applied Sciences North Western Switzerland (FHNW), 5210 Windisch, Switzerland;\\ \scriptsize{mohamad.dia@fhnw.ch}, \scriptsize{martin.melchior@fhnw.ch}}
\affil{$^2$Laboratory of Astrophysics, Ecole Polytechnique F\'ed\'erale de Lausanne (EPFL),  Observatoire de Sauverny,1290 Versoix, Switzerland;\\ \scriptsize{elodie.savary@epfl.ch}, \scriptsize{frederic.courbin@epfl.ch}}

\paperauthor{Mohamad~Dia}{mohamad.dia@fhnw.ch}{0000-0003-1028-787X}{University of Applied Sciences North Western Switzerland (FHNW)}{Institute for Data Science}{Windisch}{AG}{5210}{Switzerland}
\paperauthor{Elodie ~Savary}{elodie.savary@epfl.ch}{}{Ecole Polytechnique F\'ed\'erale de Lausanne (EPFL)}{Observatoire de Sauverny}{Versoix}{GE}{1290}{Switzerland}
\paperauthor{Martin~Melchior}{martin.melchior@fhnw.ch}{}{University of Applied Sciences North Western Switzerland (FHNW)}{Institute for Data Science}{Windisch}{AG}{5210}{Switzerland}
\paperauthor{Frederic~Courbin}{frederic.courbin@epfl.ch}{}{Ecole Polytechnique F\'ed\'erale de Lausanne (EPFL)}{Observatoire de Sauverny}{Versoix}{GE}{1290}{Switzerland}



\begin{abstract}
In this work, we provide an efficient and realistic data-driven approach to simulate astronomical images using deep generative models from machine learning. Our solution is based on a variant of the generative adversarial network (GAN) with progressive training methodology and Wasserstein cost function. The proposed solution generates naturalistic images of galaxies that show complex structures and high diversity, which suggests that data-driven simulations using machine learning can replace many of the expensive model-driven methods used in astronomical data processing.
\end{abstract}



\section{Introduction}
Investigating the reasons behind the accelerated expansion of the universe is one of the main challenges in astronomy and modern cosmology. Future space missions, such as \emph{Euclid}, will provide images of billions of galaxies in order to investigate the so-called \emph{dark matter} and probe the geometry of the universe through the \emph{gravitational lensing} effect. Due to the very large-scale of data provided by such missions, automated algorithms are needed for measurement and detection purposes.
The training and calibration of such algorithms require simulated, or synthetic, images of galaxies that mimic the real observations and exhibit real morphologies.



In the case of weak lensing for instance, the accuracy of the shape measurement algorithms is very sensitive to any statistical bias induced by the Point Spread Function (PSF). Therefore, simulated images of galaxies with known ground-truth lensing are required to calibrate and detect any potential bias in the ensemble averages. Moreover, the training of automated strong lensing detector, such as deep learning architectures \citep{lensFinding2019}, requires simulated images in order to mitigate class imbalance and avoid false-positive type of error in the current datasets.


\section{Model-Driven v.s. Data-Driven  Galaxy Image Simulation}
The current approaches to simulate images of galaxies in the cosmology literature are mostly model-driven, or rule-based, approaches. These might involve the fitting of parametric analytic profiles (size, ellipticity, brightness, etc.) to the observed galaxies. This approach is usually unable to reproduce all the complex morphologies. An alternative, more expensive and often infeasible, model-driven approach is to start with high-quality galaxy images as the input of the simulation pipeline followed by a model that reproduces all the data acquisition effects  \citep{galsim2015}.

Recently, several data-driven approaches have been investigated in order to generate synthetic images of galaxies via generative models used in machine learning \citep{celeste2015,EnablingDE2016}, mainly variational autoencoder (VAE) \citep{Kingma2013AutoEncodingVB} and generative adversarial network (GAN) \citep{GAN2014}. Such approaches have shown some promising preliminary results in generating galaxy images. Following this data-driven approach, and motivated by the success and recent impressive improvements in GANs, we have further investigated the use of such architecture in generating galaxy images.

\section{Generative Adversarial Network}
Unlike most of the generative models used in machine learning, GAN represents a novel approach that learns how to sample from the data distribution without explicitly tracking the parameters of the probability distribution function via traditional maximum likelihood estimation. The GAN architecture consists of two neural networks that compete against each other in a two-player minimax game. The first network is the ``generator'' that is responsible of generating the data, while the second network is the ``discriminator'' that represents the adversarial loss function. Despite its elegant mathematical formulation and the theoretical guarantees provided by a non-parametric analysis, the initial GAN architecture suffered from some practical implementation problems.

After the invention of GAN in 2014, a plethora of work have been done to improve the training (in terms of convergence and stability) and to obtain more realistic generated data (in terms of quality and diversity). Most of this effort was made towards improving the cost function and stabilizing the training methodology, which has recently lead to unprecedented results in generating synthetic images. Based on these recent advances, we have investigated variants of GAN that use the Wasserstein distance \citep{Wasserstein2017} and the progressive training \citep{karras2018progressive} on galaxy images provided by the Galaxy-Zoo dataset \citep{galaxyZoo}.

\section{Proposed Architecture}
Following \citep{karras2018progressive}, we employ blocks of convolutional layers to progressively build the generator and the discriminator as mirror images of each other (see Table \ref{table:architecture}). Intuitively speaking, training a small network to generate low-resolution images that capture the large-scale structure of the galaxies is an easier task than directly training a full network to generate high-resolution images with fine details. Hence, we start by training the network to generate low-resolution images ($4 \times 4$), we then progressively increase the resolution, in $4$ steps until $64 \times 64$ resolution, by \emph{smoothly} and synchronously adding blocks of convolutional layers to both the generator and discriminator. For the generator, each progress block is preceded by an up-sampling operation while a down-sampling operation follows each progress block in the discriminator.\footnote{One can also use fractionally-strided and strided convolution respectively.} Such methodology leads to a more stable and faster training.
\begin{table}[!ht]
\caption{Blocks of convolutional layers added progressively for both the generator and the discriminator.}
\label{table:architecture}
\smallskip
\scriptsize
\begin{minipage}[b]{0.45\linewidth}\centering
\begin{tabular}{cc}  
\tableline
 \textbf{Generator}  & Output Dimensions  \\
\tableline
\underline{Latent Space}  \\
Input latent vector & $1 \times 1 \times 256 $\\
Conv $4 \times 4$ & $4 \times 4 \times 256$ \\
Conv $3 \times 3$ & $4 \times 4 \times 256$ \\
\tableline
\underline{$1^{st}$ Progress}  \\
Conv $3 \times 3$ & $8 \times 8 \times 128$ \\
Conv $3 \times 3$ & $8 \times 8 \times 128$ \\
\tableline\
\underline{$2^{nd}$ Progress}  \\
Conv $3 \times 3$ & $16 \times 16 \times 64$ \\
Conv $3 \times 3$ & $16 \times 16 \times 64$ \\
\tableline\
\underline{$3^{rd}$ Progress}  \\
Conv $3 \times 3$ & $32 \times 32 \times 32$ \\
Conv $3 \times 3$ & $32 \times 32 \times 32$ \\
\tableline\
\underline{$4^{th}$ Progress}  \\
Conv $3 \times 3$ & $64 \times 64 \times 16$ \\
Conv $3 \times 3$ & $64 \times 64 \times 16$ \\
\tableline\
\underline{RGB Extraction}  \\
Conv $1 \times 1$ & $64 \times 64 \times 3$ \\
\\
\tableline\
\end{tabular}
\end{minipage}
\hspace{0.5cm}
\begin{minipage}[b]{0.45\linewidth}
\smallskip
\centering
\begin{tabular}{cc}  
\tableline
 \textbf{Discriminator}  & Output Dimensions  \\
\tableline
\underline{RGB Reading}  \\
Input image & $64 \times 64 \times 3 $\\
Conv $1 \times 1$ & $64 \times 64 \times 16$ \\
\tableline
\underline{$4^{th}$ Progress}  \\
Conv $3 \times 3$ & $64 \times 64 \times 16$ \\
Conv $3 \times 3$ & $64 \times 64 \times 32$ \\
\tableline\
\underline{$3^{rd}$ Progress}  \\
Conv $3 \times 3$ & $32 \times 32 \times 32$ \\
Conv $3 \times 3$ & $32 \times 32 \times 64$ \\
\tableline\
\underline{$2^{nd}$ Progress}  \\
Conv $3 \times 3$ & $16 \times 16 \times 64$ \\
Conv $3 \times 3$ & $16 \times 16 \times 128$ \\
\tableline\
\underline{$1^{st}$ Progress}  \\
Conv $3 \times 3$ & $8 \times 8 \times 128$ \\
Conv $3 \times 3$ & $8 \times 8 \times 256$ \\
\tableline\
\underline{Cost Calculation}  \\
Conv $3 \times 3$ & $4 \times 4 \times 256$ \\
Conv $4 \times 4$ & $1 \times 1 \times 256$ \\
Conv $1 \times 1$ & $1 \times 1 \times 1$ \\
\tableline\
\end{tabular}
\end{minipage}
\end{table}

Moreover, the Wasserstein distance with gradient penalty \citep{Wasserstein2017} is used as a cost function to mitigate the gradient problems. Furthermore, various normalization techniques are used to avoid the unhealthy competition between the generator and discriminator. In particular, we use ``weight scaling'' and ``pixelwise feature normalization'' as done in \citep{He2015,AlexNet2012}. In addition to that, the ``mini-batch standard deviation'' \citep{karras2018progressive} is computed and incorporated in the cost function in order to favor diversity in the synthetic data. 

\subsection{Results}
Our architecture is implemented in Python using PyTorch library and trained on a GPU system. The dataset is made of 6157 images of galaxies in RGB format. The images were centered at $64 \times 64$ resolution, normalized, and augmented using standard data augmentation techniques. A batch size $16$ was used with $8$ data loading workers.

The training was performed over a total of $100$ epochs and lasted less than $24$ hours. During the first $40$ epochs of training, the generator and discriminator were competing to reach the minimax equilibrium and the performance was fluctuating (in terms of their loss functions). The performance stabilized after that while the image quality continued to improve. After training, the discriminator, which plays the role of an adaptive loss function, is detached from the architecture and dismissed. The generator is then able to generate galaxy images starting from a latent space made of $256$ standard Gaussian i.i.d. random variables.


%
\begin{figure}[t!]
\centering
\includegraphics[width=0.25\textwidth, height=163pt, trim={0pt 3 3 0},clip]{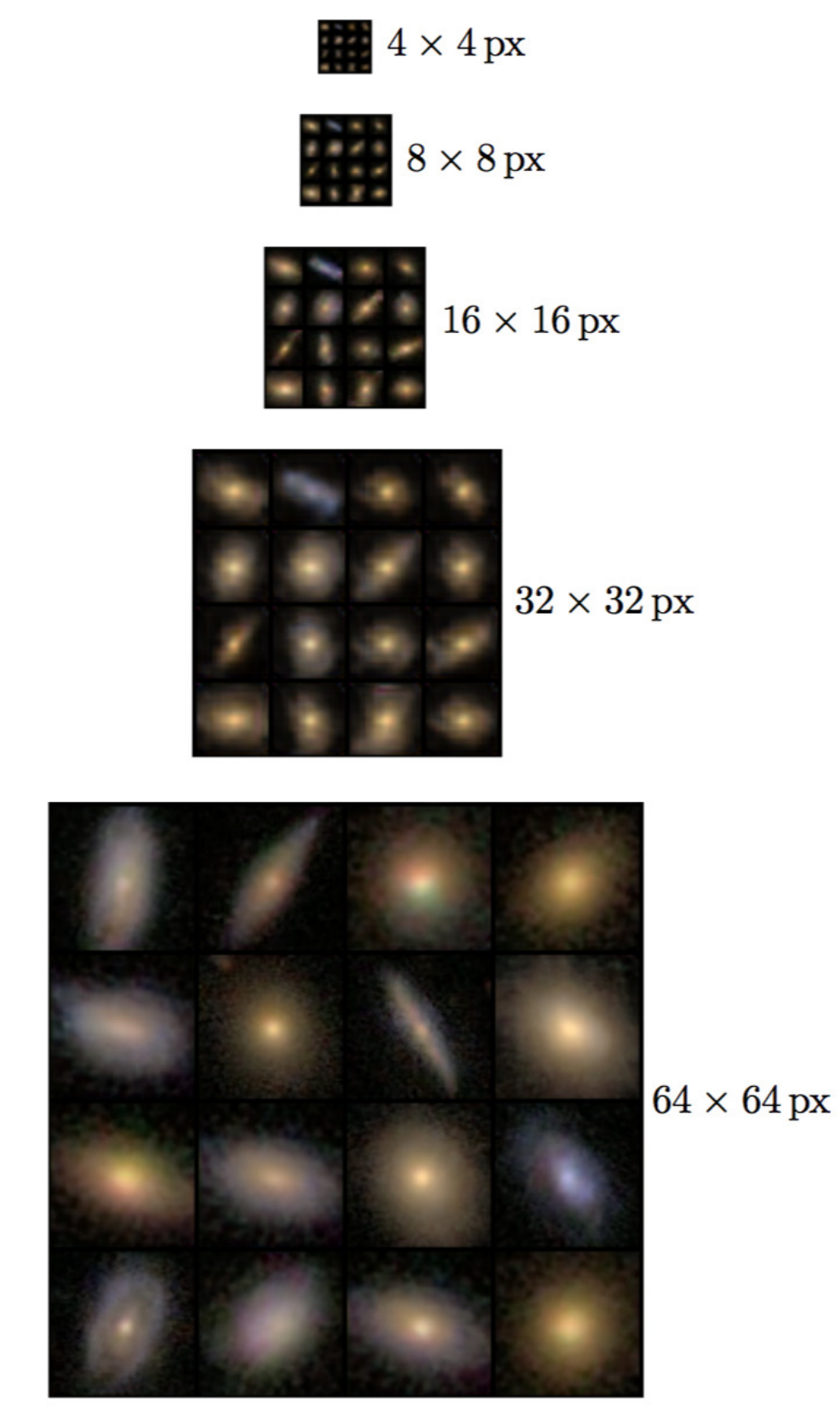}
\centering
\includegraphics[width=0.59\textwidth, height=163pt, trim={-200pt 3 3 0},clip]{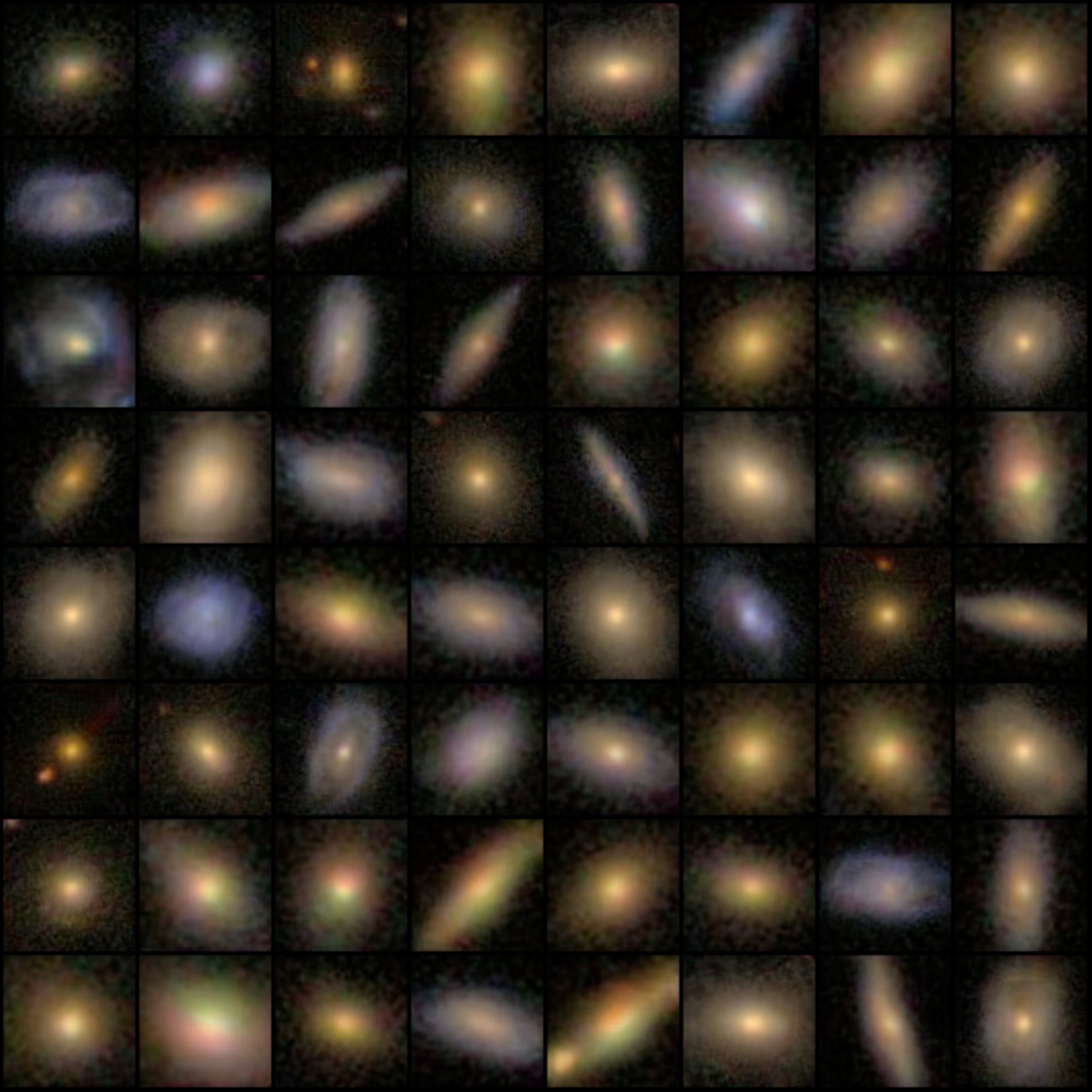}
\caption{\emph{Left:} Progressive increase of resolution in four steps. \emph{Right:} Diverse set of simulated images obtained by the proposed GAN architecture.}
\label{fig:galaxies}
\end{figure}

By changing the latent vector, we were able to obtain very diverse and high quality images of galaxies showing complex structures and morphologies (e.g. arm and disk features). Furthermore, the simulated images exhibited realistic effects (e.g. companion stars) as shown in Figure \ref{fig:galaxies}. 

\section{Future Work}
We are planning to investigate the latent space of our GAN model in order to gain insight on the effect of each latent variable on the galaxies morphology. This will provide us with more control on the generation task and will permit to interpolate between the variables and perform latent space arithmetics. Furthermore, we are planning to incorporate the labels of the galaxies, when available, in a supervised or semi-supervised approach using variants of "Conditional GAN" architectures \citep{Odena2016ConditionalIS} in order to improve the quality of the generated images and guide the generator.

\acknowledgements M. Dia and E. Savary would like to acknowledge funding from the SNSF (grant number 173716).

\bibliography{P3-14.bib}  


\end{document}